\documentclass[11pt]{article}

\usepackage{arxiv}

\usepackage[utf8]{inputenc}
\usepackage[T1]{fontenc}
\usepackage{hyperref}
\usepackage{url}
\usepackage{booktabs}
\usepackage{amsmath}
\usepackage{amsfonts}
\usepackage{amsthm}
\newtheorem{claim}{Claim}
\usepackage{microtype}
\usepackage{graphicx}
\usepackage[ruled,vlined]{algorithm2e}

\newcommand{\blfootnote}[1]{%
  \begingroup\renewcommand\thefootnote{}\footnote{#1}\addtocounter{footnote}{-1}\endgroup}

\title{Can I Buy Your KV Cache?}

\author{%
  Luoyuan Zhang\\
  Harbin Institute of Technology, Shenzhen (HITSZ)\\
  \texttt{zly.idleness@gmail.com}
}

\begin{document}
\maketitle

\begin{abstract}
Right now, across the world, AI agents are repeating the same absurd act: to read
one document, they each recompute it from scratch. Every agent re-runs
\emph{prefill}, the most compute-intensive step a large model takes, over
identical text, only to rebuild a key--value (KV) cache \emph{identical} to the
one the agent before it just built. The same answer, computed a million times. We
make a proposal that is almost offensively simple: \emph{compute it once.} Let a
publisher precompute a document's KV cache, and let every other agent \emph{buy}
the right to load it and skip prefill. It works, and it is \emph{token-exact}:
loading a precomputed KV and continuing matches prefilling from scratch ($24/24$
greedy tokens, and at the logits level), with \emph{no} accuracy cost. On
Qwen3-4B, reuse is $\mathbf{9}$--$\mathbf{50\times}$ cheaper in compute than
prefill, and the gap \emph{widens} with length (prefill's attention scales with
$L^2$), so a single reuse already pays it back.
Then the part that matters: \emph{where the KV lives}. \emph{Shipping} it fails,
because KV is nearly incompressible, so per-load egress costs more than the prefill it
saves. \emph{Hosting} it provider-side, exactly as production prompt-caching works,
removes egress entirely. The size of the prize is set by our \emph{measured}
compute saving: serving one hot $3774$-token document to $80$M agents costs
$\sim$\$1.5M to re-prefill but only $\sim$\$0.03M of reuse compute ($49.7\times$
less). The $0.1\times$ cache-read tariff APIs charge passes a $10\times$ discount
to users while sitting \emph{inside} this measured envelope, so the $10\times$
is a \emph{floor} that the measured $\sim$$50\times$ compute saving clears, and the
gap to the physical $\sim$$50\times$ is provider margin: millions of dollars per
popular document. We frame the resulting
agent-native \emph{prefill CDN} and leave lossless KV compression and a
cross-party payment layer as the open problems.
\end{abstract}

\blfootnote{Code and data: \url{https://github.com/zly-idleness/kvstore}.}

\section{Introduction}
Serving a long context to a large language model (LLM) proceeds in two phases.
\emph{Prefill} runs a forward pass over all $L$ input tokens to populate the
key--value (KV) cache; \emph{decode} then generates output tokens one at a time,
each attending to the cached keys and values. Prefill is compute-bound and its
cost grows (super-)linearly in $L$: the dense projections contribute work
proportional to $L$, and self-attention contributes a term proportional to
$L^2$. For long inputs, prefill dominates both latency (time-to-first-token) and
the compute bill.

A defining feature of retrieval-augmented and agentic workloads is that the
\emph{same content is read over and over}. A popular knowledge-base article, a
product manual, a regulatory filing, or a market report is consulted by
thousands of independent requests, sessions, or autonomous agents. Under the
standard serving model, each consumer re-prefills the identical text, repeating
exactly the same computation. As the number of consumers grows, the aggregate
waste grows with it: $N$ readers of one document perform $N$ identical prefills.

We take a simple position: \textbf{when a piece of content is public and read by
many consumers, its KV cache is the natural unit of reuse.} Compute the KV once,
persist it as a portable \emph{artifact}, and let downstream consumers
\emph{load} it and continue generating, paying a cheap load instead of a full
prefill. The closest familiar analogue is a content delivery network (CDN), with
one twist. A CDN caches content \emph{bytes} and amortizes \emph{transmission}
across many downloads; here we cache the content's \emph{KV} and amortize
\emph{computation} across many reads. We therefore call the resulting layer a
\emph{prefill CDN}: a service in which content publishers ship precomputed KV
artifacts and consuming agents pay per load, skipping prefill entirely.

\paragraph{Three objections, and why each weakens in our regime.}
Three objections arise immediately. First, the KV cache carries \emph{no
information} beyond the text: it is a deterministic function of the input under a
fixed model, so receiving it is, informationally, no different from receiving the
text. Second, it is \emph{large}: in our measurements a few-thousand-token
context produces an artifact of hundreds of megabytes to over a gigabyte. Third,
it is \emph{model-bound}: a cache produced by one model and precision is
meaningless to another. Each objection, however, weakens precisely in the regime
we target. ``No extra information'' is irrelevant when the consumer's goal is to
\emph{skip computation}, not to learn something new. Size is amortized when the
same artifact serves many reads, exactly as a CDN amortizes the cost of an edge
replica; and the artifact can be compressed (Section~\ref{sec:exp}). Model
binding is acceptable inside an ecosystem that designates one or a few standard
models, which is the natural deployment context for a shared content layer.

\paragraph{Scope.} The space of KV reuse ranges from reusing a shared prefix to
fusing the caches of several independently-prefilled fragments. We deliberately
study the simplest, \emph{safe} form: a document treated as a shared prefix,
with continuations appended after it. This form is attractive because it is
\emph{exactly} correct (the loaded KV is bit-for-bit what a from-scratch
prefill would have produced), so any cost saving comes for free, without an
accuracy trade-off. Cross-fragment fusion, which requires selective recomputation
to repair lost cross-attention, is the subject of prior work that we build on
rather than duplicate.

\paragraph{Contributions.} This paper provides a minimal, fully reproducible
study of the core question: \emph{how does the amortized per-call cost of KV
reuse compare to repeated prefill, and how does it scale with the number of
reuses?} Concretely:
\begin{itemize}
  \item \textbf{Correctness.} We show that loading a precomputed KV cache and
  continuing is \emph{token-exact} with prefilling the same context from
  scratch under greedy decoding, and we verify it both end-to-end (token match)
  and at the logits level (Section~\ref{sec:method}, Section~\ref{sec:exp}).
  \item \textbf{Cost characterization.} On Qwen3-4B we measure prefill compute
  vs.\ the compute of reusing a resident KV (one continuation step) across five
  context lengths ($255$--$3774$ tokens), finding reuse $9$--$50\times$ cheaper,
  with the gap widening with length (Section~\ref{sec:exp}).
  \item \textbf{Scaling and break-even.} We give a cost model for amortized
  per-call cost versus reuse count $N$ and show the break-even is essentially
  immediate ($N^\star\!\approx\!1$), so reuse pays off by the second read
  (Section~\ref{sec:cost}, Section~\ref{sec:exp}).
  \item \textbf{Size--exactness trade-off.} We quantify the cost of shrinking the
  artifact: int8 quantization halves it but breaks token-exact equivalence,
  delineating when lossless persistence is required (Section~\ref{sec:exp}).
  \item \textbf{A market framing.} We articulate the prefill-CDN layer (publishers
  precompute and host KV, agents pay per load over an agent-payment
  rail) and identify which technical pieces exist and which are open
  (Section~\ref{sec:disc}).
\end{itemize}
An open-source implementation (KV save/load, correctness checks, benchmarks)
accompanies the paper.

\section{Background}
\label{sec:bg}
\paragraph{Prefill and the KV cache.} A decoder-only transformer with $h$ layers
maintains, per layer, key and value tensors of shape
$(\text{batch},\,\text{heads},\,\text{seq},\,\text{head\_dim})$. Prefilling a
context of length $L$ fills these tensors for all $L$ positions. The cache size
is therefore linear in $L$: for our model (Qwen3-4B, $36$ layers, fp16) a
context of $\sim$$3.8$K tokens yields an artifact of $\sim$$557$\,MB
(Section~\ref{sec:exp}). Decode steps read this cache and append one position at
a time.

\paragraph{Prefix caching.} Production servers already exploit one form of reuse:
if two requests share a token-identical prefix, the prefix's KV can be computed
once and reused, e.g.\ via RadixAttention-style prefix trees. This is an
\emph{in-system} optimization: the cache lives in the server's memory and is
reused across that server's concurrent requests. Our framing differs in two
ways: the cache is (i) \emph{persisted} as a first-class artifact and (ii)
potentially \emph{shared across parties}, i.e.\ produced by a publisher and
consumed by independent agents.

\section{Related Work}
\label{sec:related}
Our work sits at the intersection of three lines: reusing precomputed KV across
queries, persisting/transferring KV as state, and compressing KV. We summarize
each and state precisely how our contribution differs.

\paragraph{Reusing precomputed KV across queries.}
The most direct precedents precompute KV for content chunks and reuse them across
different downstream questions. \textbf{CacheBlend}~\cite{cacheblend} observes
that naively concatenating independently-prefilled chunk caches loses the
cross-attention between chunks, degrading quality; it restores quality by
\emph{selectively recomputing} a small fraction of tokens (those whose attention
is most affected), recovering most of the speedup of full reuse while keeping
accuracy. \textbf{Cache-Craft}~\cite{cachecraft} manages a pool of reusable
``chunk-caches'' for retrieval-augmented generation, deciding which chunks to
cache and when to recompute, and reports large reductions in redundant compute
relative to prefix caching. \textbf{CacheClip}~\cite{cacheclip} uses small
auxiliary models to guide which tokens to recompute when reusing KV, retaining
most of full-attention quality. These methods target the \emph{hard} case, fusing caches of fragments that did
not attend to one another, and accept a controlled accuracy cost in exchange for
position-independent reuse. Our study is
deliberately complementary and narrower: we examine the \emph{exact} case (a
single shared prefix), where no recomputation is needed and reuse is token-exact,
and we focus on the \emph{cost scaling with the number of reuses} and on the
framing of KV as a publishable artifact, rather than proposing a new fusion
mechanism. CacheBlend-style fusion is orthogonal and could sit underneath a
prefill CDN to broaden what is reusable.

\paragraph{KV as persisted or transferable state.}
A second line treats the KV cache as state that lives outside a single GPU's
memory. \textbf{MatKV}~\cite{matkv} precomputes KV for RAG objects and
\emph{materializes them in flash storage}, trading compute for cheap, fast,
power-efficient storage and reporting halved inference time and energy.
\textbf{InstInfer}~\cite{instinfer} pushes the idea further into the storage
device, offloading attention computation and KV management to computational
storage drives for long-context inference. In disaggregated serving, prefill and
decode run on different machines, so the KV cache becomes a \emph{network
payload} routed across the datacenter: \textbf{NetKV}~\cite{netkv} selects decode
instances with awareness of the network cost of moving the cache, and
\textbf{ConServe}~\cite{conserve} schedules at conversation granularity,
transferring a conversation's KV once to a pinned decoder. Collectively, these
works establish that KV can be stored to disk and moved across machines, the
mechanical feasibility our artifact framing relies on. What is missing, and what
we add, is the \emph{economic} layer: KV as a cross-party, payable good shipped
by a publisher to many independent consumers, and the corresponding cost-scaling
argument.

\paragraph{KV compression.}
The size of the artifact is the main obstacle to distribution, and a rich body of
work attacks it. \textbf{SIFT}~\cite{sift} avoids storing full KV altogether,
keeping only the indices of high-attention locations as compact bit vectors that
are orders of magnitude smaller, then recomputing on demand.
\textbf{SpectrumKV}~\cite{spectrumkv} treats KV transfer in disaggregated serving
as a precision-allocation problem, assigning per-token bit-widths
(fp16/int8/int4) to minimize transfer volume at a given quality. Our int8
experiment (Section~\ref{sec:exp}) is a simple instance of this trade-off and
illustrates why such methods are needed: uniform int8 already breaks token-exact
equivalence, so non-uniform / selective schemes are necessary to push size down
without sacrificing reproducibility. Compression is fully complementary to our
study: it improves the economics of the prefill CDN without changing its
structure.

\paragraph{Positioning.} In one sentence: prior work makes KV reuse
\emph{possible} (fusion), \emph{storable} (flash/CSD), \emph{movable}
(disaggregation), and \emph{small} (compression); we ask what happens when KV is
treated as a \emph{published artifact reused by many parties}, characterize the
cost scaling that makes this attractive, and verify the exact-reuse case end to
end.

\section{Method}
\label{sec:method}
We formalize the two operations, producing an artifact and consuming it,
and then argue correctness for the shared-prefix case.

\paragraph{Setup and notation.} Fix a decoder-only model $f_\theta$ with $h$
layers. For a token sequence $x_{1:L}$, prefill produces a cache
$\mathcal{K}(x_{1:L}) = \{(K^{(\ell)}, V^{(\ell)})\}_{\ell=1}^{h}$, where
$K^{(\ell)}, V^{(\ell)}$ hold the per-position keys/values at layer $\ell$ for
positions $1\dots L$. Crucially, under a fixed model and deterministic kernels,
$\mathcal{K}$ is a \emph{deterministic function} of $x_{1:L}$: the entry for
position $i$ depends only on $x_{1:i}$ (causal attention), not on any tokens that
come later.

\subsection{Producing an artifact (publisher)}
Given a context document $c = x_{1:L}$, the publisher runs one forward pass to
obtain $\mathcal{K}(c)$, then serializes the per-layer key/value tensors to disk
together with metadata $m=(\text{model id},\ \text{dtype},\ L)$
(Algorithm~\ref{alg:save}). This is a one-time cost. The metadata is essential:
the artifact is only valid for the exact $(\text{model},\text{dtype})$ pair it
was produced with, and the consumer must reject mismatches.

\begin{algorithm}[t]
\caption{\textsc{SaveKV} (publisher, one-time)}
\label{alg:save}
\KwIn{context $c=x_{1:L}$, model $f_\theta$}
\KwOut{artifact $a$ on disk}
$\mathcal{K} \leftarrow f_\theta.\textsc{forward}(c,\ \texttt{use\_cache=True})$\;
\For{$\ell = 1$ \KwTo $h$}{
  store $(K^{(\ell)}, V^{(\ell)})$ to CPU tensors\;
}
$a \leftarrow \textsc{serialize}(\{(K^{(\ell)},V^{(\ell)})\},\ m{=}(\text{model},\text{dtype},L))$\;
\Return $a$\;
\end{algorithm}

\subsection{Consuming an artifact (agent)}
To answer a query, the consumer loads $\mathcal{K}(c)$, reconstructs the model's
cache object, and decodes the continuation $q$ \emph{without re-prefilling} $c$
(Algorithm~\ref{alg:load}). Two details make this exact. (i) \emph{Positions}:
the continuation tokens must be assigned position ids $L, L{+}1, \dots$ so that
positional encodings (e.g.\ RoPE) match what a from-scratch pass would use. (ii)
\emph{Attention mask}: the mask must span the full $L + |q|$ positions so the
continuation attends to the entire cached context. Omitting either reproduces the
``shifted'' failure mode in which outputs diverge.

\begin{algorithm}[t]
\caption{\textsc{GenerateFromKV} (agent, per query)}
\label{alg:load}
\KwIn{artifact $a$, continuation tokens $q_{1:M}$, length $n$}
\KwOut{generated tokens $y_{1:n}$}
$(\mathcal{K}, m) \leftarrow \textsc{deserialize}(a)$\;
\textbf{assert} $m.\text{model}, m.\text{dtype}$ match the running model\;
$\text{cur} \leftarrow q_{1:M}$;\quad $\text{start} \leftarrow m.L$;\quad $y \leftarrow [\,]$\;
\For{$t = 1$ \KwTo $n$}{
  $S \leftarrow \text{start} + |\text{cur}|$\;
  $\text{pos} \leftarrow [\text{start}, \dots, S{-}1]$;\quad
  $\text{mask} \leftarrow \mathbf{1}^{S}$\;
  $o \leftarrow f_\theta(\text{cur},\ \mathcal{K},\ \text{pos},\ \text{mask})$\;
  $y_t \leftarrow \arg\max o.\text{logits}[-1]$;\quad append to $y$\;
  $\mathcal{K} \leftarrow o.\text{cache}$;\quad $\text{cur}\leftarrow [y_t]$;\quad $\text{start}\leftarrow S$\;
}
\Return $y$\;
\end{algorithm}

\subsection{Correctness of exact prefix reuse}
\label{sec:correct}
\begin{claim}
Let $c=x_{1:L}$ and continuation $q$. Under a fixed model with deterministic
kernels and greedy decoding, \textsc{GenerateFromKV}$(\textsc{SaveKV}(c), q, n)$
produces the same tokens as prefilling $c\,\Vert\,q$ from scratch and
greedy-decoding $n$ tokens.
\end{claim}
\emph{Argument.} Because attention is causal, the cache entries for positions
$1\dots L$ are identical whether produced while prefilling $c$ alone or while
prefilling $c\,\Vert\,q$: they never depend on tokens after position $L$.
Serialization/deserialization is value-preserving. Given correct position ids and
a full attention mask, the forward pass over $q$ attends to exactly the same keys
and values, at the same relative positions, as the from-scratch pass. Hence the
logits at every continuation position match, and greedy \texttt{argmax} yields
identical tokens. In practice, non-associative floating-point and nondeterministic
GPU/MPS kernels perturb logits by a tiny amount; we therefore compare
\emph{argmax} tokens (the decision-relevant quantity) and also report the maximum
absolute logit difference as evidence the perturbation is negligible
(Section~\ref{sec:exp}).

\paragraph{Why only prefixes.} The argument relies on the cached positions being
a causal \emph{prefix} of the full sequence. If instead one stitches together
caches of fragments $c_1, c_2$ that were prefilled independently, the entries for
$c_2$ never attended to $c_1$, so cross-attention is missing and exactness no
longer holds. This is precisely the regime that fusion methods
\cite{cacheblend,cachecraft} repair via selective recomputation. We restrict to
prefixes to keep reuse free of any accuracy cost.

\section{Cost Model}
\label{sec:cost}
For one document, let $C_{\text{prefill}}$ be the cost to compute its KV and
$C_{\text{load}}$ the cost to load the precomputed KV. Over $N$ reuses:
\begin{align}
\text{from-scratch:}\quad & \text{total} = N\,C_{\text{prefill}},
  & \text{per-call} = C_{\text{prefill}} \\
\text{KV-reuse:}\quad & \text{total} = C_{\text{prefill}} + N\,C_{\text{load}},
  & \text{per-call} = \frac{C_{\text{prefill}}}{N} + C_{\text{load}}.
\end{align}
The from-scratch per-call cost is flat; the KV-reuse per-call cost decreases with
$N$ toward a floor of $C_{\text{load}}$. The break-even reuse count is
\begin{equation}
N^\star = \frac{C_{\text{prefill}}}{C_{\text{prefill}} - C_{\text{load}}}.
\end{equation}
When $C_{\text{load}} \ll C_{\text{prefill}}$, $N^\star \to 1$: reuse wins almost
immediately, and the absolute saving $N(C_{\text{prefill}}-C_{\text{load}})$
grows with both the reuse count and the document length (longer documents have
larger $C_{\text{prefill}}$).

\paragraph{A storage-aware model.} The latency-only model above ignores the cost
of \emph{storing} and \emph{moving} the artifact, which a real prefill CDN must
pay. Let $s$ be the artifact size (bytes), $\sigma$ a storage price (per byte,
per unit time over a hosting horizon $T$), and $\beta$ a transfer cost (per byte,
charged once per load). Then
\begin{equation}
\text{total}_{\text{KV}}(N) = C_{\text{prefill}} + \sigma s T + N\,(C_{\text{load}} + \beta s),
\end{equation}
so the effective per-load floor rises from $C_{\text{load}}$ to
$C_{\text{load}} + \beta s + \sigma s T / N$. Two consequences follow. (i)
Because $s$ scales with context length while $C_{\text{prefill}}$ scales
(super-)linearly with length and is multiplied by the (large) per-token compute
cost, the saving still grows with length, but the transfer term $\beta s$ caps
how cheap a load can get, which is exactly why artifact \emph{compression}
\cite{sift,spectrumkv} matters for the economics as well as for storage. (ii) The
amortized storage term $\sigma s T / N$ vanishes as $N$ grows, so the model again
rewards heavy reuse. In short, the latency model is the $\sigma{=}\beta{=}0$
special case; compression and high reuse are what keep the storage-aware floor
low.

\section{Experiments}
\label{sec:exp}
\paragraph{Setup.} Qwen3-4B in float16 on Apple M1 Pro (MPS). We use float16 (not
quantization) for the correctness check, since quantization can break token-exact
equivalence.

\paragraph{Correctness.} Loading a precomputed KV cache and continuing is
\textbf{token-exact} with prefilling from scratch: under greedy decoding we
observe a $24/24$ token match. A logits-level check on the next-token prediction
gives identical argmax and a maximum absolute logit difference of $0.02$,
consistent with floating-point nondeterminism on MPS rather than a semantic
difference.

\paragraph{Compute cost across lengths.} We compare the \emph{compute} of two
operations, with the context KV resident in memory in both the hosted case (no
disk I/O, no re-prefill): (a) a full prefill of the $L$-token context, versus
(b) one continuation step that attends over the already-cached context.
Table~\ref{tab:cost} reports the median of $5$ runs. Reuse is
$\mathbf{8.6}$--$\mathbf{49.7\times}$ cheaper in compute, and the advantage grows
with length: prefill's attention term scales with $L^2$ while a single
continuation step grows far more slowly, so the ratio widens monotonically. We
deliberately report a regime free of the device memory-pressure artifact
discussed in Limitations (a $7.5$K-token prefill on this device inflates
super-linearly and would overstate the ratio); the numbers below are the
honest, memory-stable range.

\begin{table}[t]
\centering
\caption{Compute cost of full prefill vs.\ one continuation step over a
\emph{resident} KV cache (Qwen3-4B, fp16, Apple M1 Pro / MPS; median of $5$ runs,
disk I/O excluded). KV size is the analytic fp16 cache size. Speedup
$=C_{\text{prefill}}/C_{\text{reuse}}$.}
\label{tab:cost}
\begin{tabular}{rrrrr}
\toprule
Tokens & Prefill (s) & Reuse-step (s) & KV size (MB) & Speedup \\
\midrule
255  & 0.67  & 0.078 & 38  & 8.6$\times$ \\
485  & 1.29  & 0.093 & 72  & 13.9$\times$ \\
945  & 2.65  & 0.122 & 139 & 21.7$\times$ \\
1888 & 5.81  & 0.179 & 278 & 32.4$\times$ \\
3774 & 14.71 & 0.296 & 557 & 49.7$\times$ \\
\bottomrule
\end{tabular}
\end{table}

\begin{figure}[t]
\centering
\includegraphics[width=0.7\linewidth]{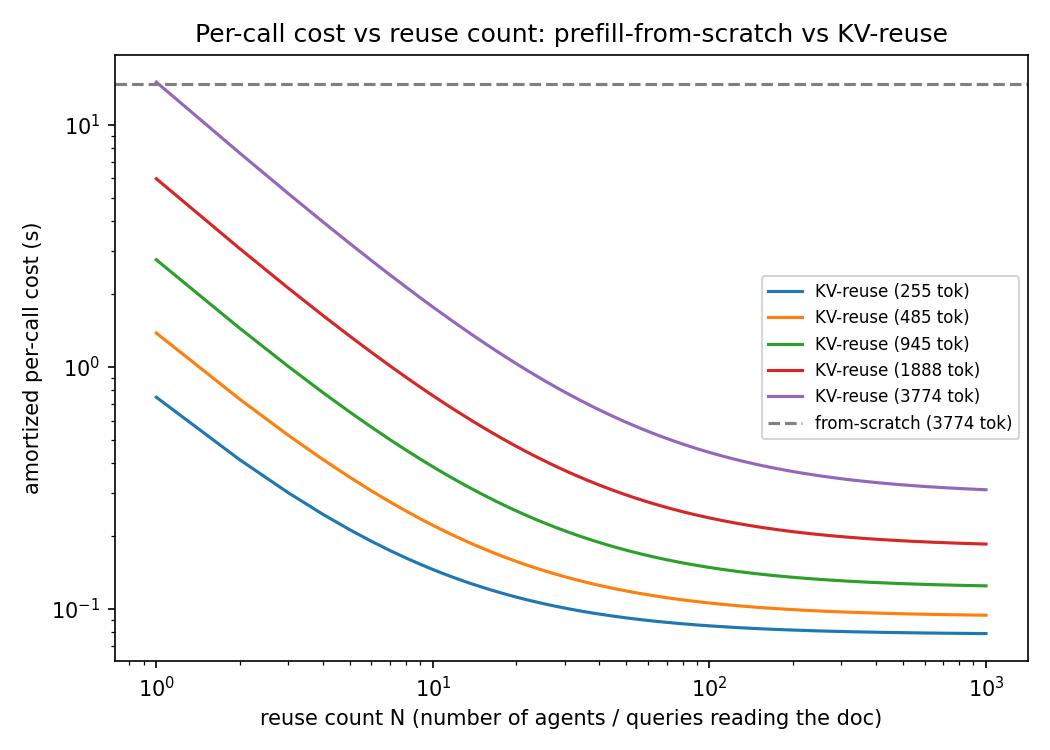}
\caption{Amortized per-call cost vs.\ reuse count $N$ (log--log). The
from-scratch cost is flat at $C_{\text{prefill}}$; KV-reuse falls as
$C_{\text{prefill}}/N + C_{\text{reuse}}$ toward a floor of $C_{\text{reuse}}$.}
\label{fig:scaling}
\end{figure}

\paragraph{Scaling.} Figure~\ref{fig:scaling} plots the amortized per-call cost
versus reuse count $N$. Because $C_{\text{reuse}} \ll C_{\text{prefill}}$, the
break-even is essentially immediate ($N^\star \in [1.02, 1.13]$): KV reuse pays
off by the second read. The absolute compute saving grows with both $N$ and
document length: at $N{=}1000$, reuse saves $\sim$$0.6{\times}10^{3}$\,s for
the 255-token document and $\sim$$1.4{\times}10^{4}$\,s for the 3774-token
document.

\paragraph{Quantizing the artifact: size vs.\ exactness.} The KV artifact is
large (Table~\ref{tab:cost}), so a natural lever is to quantize it. We store a
$692$-token context KV at fp16 and at per-tensor symmetric int8. Int8 halves the
artifact ($102.1\to51.1$\,MB, $2.0\times$) but \emph{breaks token-exact
equivalence}: greedy continuation matches the fp16 path for only $16/32$ tokens,
diverging once dequantization errors accumulate (both outputs remain fluent and
on-topic, so the \emph{quality} impact is unquantified; this is a limitation, and a
reason this is a single illustrative probe rather than a general claim). The
takeaway is qualitative: any lossy compression forfeits the bit-exactness that
makes prefix reuse free of accuracy cost, motivating selective / mixed-precision
schemes \cite{sift,spectrumkv} when reproducibility matters.

\paragraph{Artifact size.} KV size scales linearly with context length
($\sim$$0.148$\,MB/token, fp16, $36$ layers, $8$ KV heads $\times$ head dim $128$,
grouped-query attention). This
linear growth against prefill's super-linear compute is why long documents are
simultaneously the most expensive to prefill and the most rewarding to cache:
compute saved per stored byte increases with length.

\section{Discussion: a prefill CDN for the agent economy}
\label{sec:disc}
The cost structure above suggests a market layer, and one design choice decides
whether it exists: \emph{where the KV lives}. \textbf{Shipping} the artifact to
each consumer fails: KV is high-entropy and nearly incompressible losslessly,
so the large artifact must be egressed (near) in full per load; at commodity
egress ($\$0.09$/GB) the $0.54$\,GB artifact of a $3774$-token document costs
$\sim$\$0.049 to transfer, $2.6\times$ the $\sim$\$0.019 of input-priced prefill it
saves, and the gap only worsens with length, so shipping is a net loss; only
\emph{lossy} compression makes it pay (forfeiting exactness; cf.\ our int8
result). \textbf{Hosting} the KV provider-side eliminates egress entirely,
exactly how production prompt-caching works: the cache stays in the provider's
datacenter and consumers are billed a cache-read rate per load. The economic
ceiling here is set by the \emph{measured} compute saving, not by any quoted price
ratio. From Table~\ref{tab:cost}, serving a load by attending over a resident
KV costs $C_{\text{reuse}}/C_{\text{prefill}} = 1/8.6$ to $1/49.7$ of a
from-scratch prefill, the ratio widening with length. A provider therefore has
physical room to price a cache-read anywhere above its true marginal cost
$C_{\text{reuse}}$ and below the user's alternative $C_{\text{prefill}}$; the
$0.1p_{\text{in}}$ cache-read tariff that production APIs actually charge sits
\emph{inside} this measured envelope: it extracts a $10\times$ user-facing
discount while our experiments show the underlying compute is $8.6$--$49.7\times$
cheaper, so on long documents the provider keeps the gap between physical
($\sim$$50\times$) and priced ($10\times$) savings as margin. This is the answer
to the tautology objection: the $10\times$ is a \emph{floor} grounded in physics,
which our measured $8.6$--$49.7\times$ compute saving comfortably clears,
independent of how the tariff is set. What a consumer pays for, then, is the right
to reuse a hosted cache, billed below the prefill it spares them and above the
compute it actually costs, at millions of dollars of margin per popular document.

\begin{table}[t]
\centering
\caption{Serving one hot $3774$-token document (our largest \emph{measured}
point) to $N{=}80$M agents. The \emph{priced} column applies the production
$0.1p_{\text{in}}$ cache-read tariff; the \emph{physical} column applies our
\emph{measured} compute ratio $C_{\text{reuse}}/C_{\text{prefill}}{=}1/49.7$
(Table~\ref{tab:cost}). Hosting is cheaper than re-prefill on \emph{both} axes;
the gap between the two is the provider's margin and shows the $10\times$ is a
physically grounded floor. Shipping the raw artifact loses money to egress.}
\label{tab:dollars}
\begin{tabular}{lrr}
\toprule
Strategy & Serve $N{=}80$M & vs.\ re-prefill \\
\midrule
re-prefill (today)                       & \$1{,}509{,}600 & $1.0\times$ \\
ship raw fp16 (lossless, egress)         & \$3.9\text{M}    & $0.4\times$ (loses) \\
ship int8 ($2\times$, \emph{lossy})      & \$2.0\text{M}    & $0.8\times$ (loses, inexact) \\
host, \emph{priced} ($0.1p_{\text{in}}$) & \$150{,}960     & $10.0\times$ \\
\textbf{host, \emph{physical} (measured)} & \textbf{\$30{,}370} & $\mathbf{49.7\times}$ \\
\bottomrule
\end{tabular}
\end{table}

The substrate for this is mature: reuse \cite{cacheblend,cachecraft,cacheclip},
persistence \cite{matkv,instinfer}, transfer \cite{netkv,conserve}, and
compression \cite{sift,spectrumkv}; what remains open is the cross-party economic
layer (pricing, invalidation on model upgrade, provenance, fallback on miss) and
the standardization that exact, model-bound reuse implies.

\section{Limitations}
Our study is intentionally minimal, and several limitations qualify the results.
\textbf{Model binding.} Exact reuse requires the consumer to run the same model
and precision as the publisher; an artifact is undefined across models. This
constrains deployment to standardized-model ecosystems
(Section~\ref{sec:disc}).
\textbf{Prefix-only.} We study a document as a shared prefix; we do not handle
multiple retrieved fragments that must be fused, which needs selective
recomputation \cite{cacheblend} and trades exactness for flexibility.
\textbf{Artifact size.} Full-precision KV is large; lossless distribution at
scale depends on compression, and our int8 result shows naive quantization
already costs exactness.
\textbf{Single machine, one model.} All measurements are on one device
(Apple M1 Pro / MPS) and one model (Qwen3-4B, fp16). Absolute speedups are
device-specific; the \emph{qualitative} result (reuse-compute $\ll$ prefill, gap
growing with length) follows from the $L^2$ attention term and should hold
broadly, but cross-device (server-GPU) and cross-model confirmation is future
work. We explicitly \emph{exclude} the $\geq7.5$K-token regime, where prefill on
this device inflates super-linearly under memory pressure and would overstate the
ratio; the reported $9$--$50\times$ is the memory-stable range.
\textbf{Reuse-cost model.} We measure the compute of \emph{one} continuation step
($M{=}1$) over a resident cache; a consumer query of $M{>}1$ tokens pays a small
prefill over its own (short) suffix, a cost \emph{constant} in the cached length
$L$. Our reported $8.6$--$49.7\times$ is therefore a best case in $M$; the
length-scaling trend is unaffected since the added term does not grow with $L$. A
full serving system also pays cache management and (for a shipping model)
deserialization, which we report as out of scope here.
\textbf{Cost model granularity.} Our break-even uses prefill vs.\ reuse
load; a full economic model would add network-transfer and storage costs for the
artifact, which compression directly affects.

\section{Future Work}
Several directions would turn the present study into a deployable layer.
\textbf{Cross-device and cross-model validation.} Repeating the measurements on
server GPUs and across model families would separate device-specific constants
from the (expected) device-independent trend (the $L^2$-driven widening), and
extend the measurement into the long-context regime that we excluded here to
avoid the on-device memory-pressure confound.
\textbf{Compression in the loop.} Combining selective / mixed-precision KV
\cite{sift,spectrumkv} with the artifact framing would measure the true
size--exactness--cost surface, and identify the compression level that keeps
loads cheap without breaking reproducibility for auditable use.
\textbf{Fusion under the artifact.} Placing CacheBlend-style fusion
\cite{cacheblend} beneath a prefill CDN would broaden reuse from single prefixes
to multiple retrieved fragments, at a controlled accuracy cost, extending the
market from ``one document'' to ``a retrieval set.''
\textbf{The economic layer.} A concrete pricing/invalidation/provenance design
over an agent-payment rail, with a fallback-to-local-prefill policy on cache
miss, is needed to move from feasibility to a usable service. \textbf{Selective
caching policy.} Learning the length$\times$popularity threshold above which
caching pays, online, from request traces.
\textbf{Secure hosted KV.} Encryption at rest and access control for paid,
private caches, so a hosted artifact can be sold without exposing its contents or
allowing unmetered reuse.
\textbf{Beyond text: video and multimodal KV.} Long video and multimodal
contexts have especially expensive prefill and are also heavily re-watched/re-queried,
making their KV an even stronger candidate for hosted reuse; characterizing
multimodal KV size, exactness, and reuse patterns is a natural extension.
\textbf{A KV-native content layer.} Taken together, these point to content
published \emph{with} a hosted, reusable KV as a first-class object: a
``KV-native'' repository or community in which popular documents ship a ready-to-reuse
cache and consumers (agents) transact reuse rather than re-prefilling. Designing
the formats, provenance, and incentives of such a layer is an open,
systems-and-economics-spanning direction.

\section{Conclusion}
Prefill is the dominant cost of serving long contexts, and in
retrieval-augmented and agentic workloads the same context is prefilled again and
again by independent consumers. We argued for treating the context's KV cache as
a precomputed, reusable \emph{artifact}: computed once by a publisher and loaded
by many agents. We showed that, for the shared-prefix case, this reuse is
\emph{token-exact} with from-scratch prefill, removing any accuracy concern; that
reusing a resident KV is $9$--$50\times$ cheaper in compute than prefilling on Qwen3-4B, with the gap
widening with context length; that the break-even is essentially immediate, so
reuse pays off by the second read; and that compressing the artifact (e.g.\
uniform int8) trades that exactness for size. Together these results motivate an
agent-native \emph{prefill CDN}, a layer that caches computation rather than
bytes, and delineate both the regime where it pays (long, heavily-read
content) and the open problems (compression in the loop, fusion, and the
cross-party economic layer) that a real deployment must solve.

\bibliographystyle{plain}
\bibliography{refs}

\end{document}